\documentclass[letterpaper, 10 pt, conference]{ieeeconf}  % Comment this line out if you need a4paper

\IEEEoverridecommandlockouts                              % This command is only needed if 
\usepackage{graphics} % for pdf, bitmapped graphics files
\usepackage{epsfig} % for postscript graphics files
\usepackage{mathptmx} % assumes new font selection scheme installed
\usepackage{times} % assumes new font selection scheme installed
\usepackage{multirow}
\usepackage{amsmath} % assumes amsmath package installed
\usepackage{amssymb}  % assumes amsmath package installed
\usepackage{algorithmic}
\usepackage{algorithm}
\usepackage{balance}
\usepackage{soul,xcolor}
\usepackage[hidelinks, implicit=false]{hyperref}
\let\labelindent\relax
\usepackage{enumitem}
\usepackage{scalerel}
\setitemize{leftmargin=*}
\DeclareMathOperator*{\argmax}{arg\,max}
\DeclareMathOperator*{\argmin}{arg\,min}
\title{\LARGE \bf
Accessibility-Based Clustering for Efficient Learning of Locomotion Skills
}

\author{Chong Zhang$^{1*}$, Wanming Yu$^{2*}$, and Zhibin Li$^{2\dagger}$% <-this % stops a space
\thanks{*These authors contributed equally to this work.}
\thanks{$\dagger$Corresponding author. Email: {\tt\small zhibin.li@ed.ac.uk}}% <-this % stops a space
\thanks{$^{1}$Chong Zhang is with Department of Precision Instrument, Tsinghua University, Beijing, 100084 China.
\newline Email: {\tt\small chong-zh18@mails.tsinghua.edu.cn}}%
\thanks{$^{2}$Wanming Yu and Zhibin Li are with the School of Informatics, University of Edinburgh, 10 Crichton St, Edinburgh EH8 9AB, United Kingdom.
\newline Email: {\tt\small wanming.yu@ed.ac.uk}, {\tt\small zhibin.li@ed.ac.uk}}
\thanks{Accompanying video: \url{https://youtu.be/7cZMThUn0rU}}%
}

\begin{document}
\bstctlcite{IEEEexample:BSTcontrol}

\maketitle
\thispagestyle{empty}
\pagestyle{empty}

%%%%%%%%%%%%%%%%%%%%%%%%%%%%%%%%%%%%%%%%%%%%%%%%%%%%%%%%%%%%%%%%%%%%%%%%%%%%%%%%
\begin{abstract}
For model-free deep reinforcement learning of quadruped locomotion, the initialization of robot configurations is crucial for data efficiency and robustness. This work focuses on algorithmic improvements of data efficiency and robustness simultaneously through automatic discovery of initial states, which is achieved by our proposed K-Access algorithm based on accessibility metrics. Specifically, we formulated accessibility metrics to measure the difficulty of transitions between two arbitrary states, and proposed a novel K-Access algorithm for state-space clustering that automatically discovers the centroids of the static-pose clusters based on the accessibility metrics. By using the discovered centroidal static poses as the initial states, we can improve data efficiency by reducing redundant explorations, and enhance the robustness by more effective explorations from the centroids to sampled poses. Focusing on fall recovery as a very hard set of locomotion skills, we validated our method extensively using an 8-DoF quadrupedal robot \textit{Bittle}. Compared to the baselines, the learning curve of our method converges much faster, requiring only 60\% of training episodes. With our method, the robot can successfully recover to standing poses within 3 seconds in 99.4\% of the test cases. Moreover, the method can generalize to other difficult skills successfully, such as backflipping. 
\end{abstract}
%\begin{keywords}
%Reinforcement Learning, Legged Robots, Fall Recovery, Clustering.
%\end{keywords}
%%%%%%%%%%%%%%%%%%%%%%%%%%%%%%%%%%%%%%%%%%%%%%%%%%%%%%%%%%%%%%%%%%%%%%%%%%%%%%%%
\section{Introduction}
\label{sec:intro}
% para 1 fall rec 
% Alex: You don't need to change much, just say trotting etc is easy (so the exploration limitation/problem does not expose), fall recovery is more difficult because there are many possibilities.

Among robot locomotion skills, trotting and some other tasks are easy in terms of exploration, while fall recovery is difficult due to many possible robot states. Thus, the ability to recover from a fall is critical for legged robots to improve their robustness against potential failures.
% and is almost the hardest to achieve due to the large variety of states. 
% To deliver robust and resilient real-world deployment, a high success rate of recovery from falls becomes increasingly indispensable. 

For fall recovery, manually-designed joint trajectories \cite{shamsuddin2011humanoid} are laborious, and their robustness in different environments is not guaranteed. Optimization-based \cite{mordatch2012discovery} methods require a significant amount of time to obtain a feasible solution since dynamic models and complex contact situations need to be considered \cite{melon2021receding}, which makes it difficult to achieve real-time fall recovery. To overcome the limitations of these methods, model-free deep reinforcement learning (DRL) methods \cite{yang2020multi} serve as a promising alternative.

% para 2 initialization: existing 3 kinds
However, learning fall recovery via DRL suffers from hard exploration, i.e., to explore in domains with sparse, delayed, or deceptive rewards, and redundant exploration, i.e., certain regions of the state space being too frequently visited while training because of skewed data collection. Both hard exploration and redundant exploration can detriment the data efficiency and learning performance, and initial state distribution is one of the key factors that can affect the efficiency of exploration.

Currently, there are three common ways to design initial state distributions for learning fall recovery via DRL: 1) initialization from demonstrations \cite{peng2018deepmimic}, 2) initialization from random distributions \cite{reda2020learning}, and 3) initialization from predefined poses \cite{yang2020multi}. 
For initialization from demonstrations, the performance is limited to the demonstrated examples. For initialization from random distributions, it is not data efficient \cite{reda2020learning} because of the redundant exploration. Also, corner cases may suffer from insufficient exploration because of skewed data collection. As a result, the diversity of exploration is not guaranteed. For initialization from predefined poses, states that lack heuristics are very likely to be missed, and the generalization can be a problem despite high data efficiency. 
% Among common methods of learning fall recovery, random initialization has low data efficiency, while manually predefined initialization lacks of variety or robustness. 

In this work, we aim to automatically discover initial states that can help achieve high robustness while still being data efficient.
We propose to achieve this by clustering the static poses of the robot and applying centroids as initial states.

% \begin{figure}[tpb]
%   \centering
%   \includegraphics[width=88mm]{figs/fig1.pdf}
%   \caption{Levels of redundant explorations and hard exploration across different initial state distributions. Higher redundancy means lower diversity of exploration. Our proposed automatic discovery of initial states via clustering is more data efficient.}
%   \label{fig:ini_vs_exploring}
%   \vspace{-4mm}
% \end{figure}

\subsection{Related Work}
\label{subsec:related}
Regarding DRL, model-free DRL has been a commonly-used method for fall recovery and other locomotion tasks \cite{yue2020learning} \cite{hwangbo2019learning} \cite{rudin2021learning}. In contrast to model-based methods such as model predictive control \cite{neunert2018whole} \cite{lafaye2015model}, trajectory optimization \cite{chatzinikolaidis2020contact} \cite{wang2012whole} and Bayesian optimization \cite{yuan2019bayesian}, model-free methods do not require explicit knowledge of complex dynamics, and support fast online computation without mathematical optimization. The most popular model-free DRL algorithms for robot locomotion are the PPO algorithm \cite{schulman2017proximal} and the SAC algorithm \cite{haarnoja2018soft}.
 
Regarding clustering, common clustering methods have successful applications in RL, such as value function approximation \cite{xu2014clustering}, but they fail to achieve ideal clustering of diverse robot states. For centroid-based methods such as K-Means \cite{macqueen1967some} and density-based methods such as DBSCAN \cite{ester1996density}, the problem is the metric for clustering, which will be discussed later. For pattern-based methods \cite{malliaros2013clustering}, state transitions of the robot can be regarded as edges in a directed graph. However, the existing methods could not achieve the desired effect as well, which will be discussed in \ref{subsec:contribution}.

\begin{figure}[tpb]
  \centering
  \includegraphics[width=84mm]{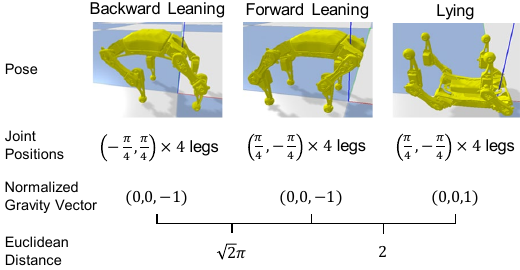}
  \caption{An erroneous case of using Euclidean distance metric. The distance between a backward leaning stance and a forward leaning stance should be smaller than that between a forward leaning stance and a lying pose.}
  \label{fig:Euclidean_fail}
  \vspace{-5mm}
\end{figure}

Regarding the metric, we usually adopt the Euclidean distance assuming orthogonal coordinates and undirected distances. However, both of the assumptions are problematic for robot poses. For the state space of static poses, we tend to use the combination of the normalized gravity vector and the joint positions \cite{yang2020multi}, and these feature dimensions are coupled. Also, it is difficult to determine the scale of the features. A failure case of the Euclidean distance is shown in Fig. \ref{fig:Euclidean_fail}. The undirected distances are inapplicable because the robot's transitions are directed. For instance, it is easy to fall from a standing pose, but difficult to recover from a lying pose.

There are also some metric learning methods, but they are time-consuming and difficult to obtain values for each start-end pose pair. In \cite{bharatheesha2014distance}, the distance metric is approximated for state-space RRTs \cite{lavalle1998rapidly}, but it is impractical to construct an RRT for each pose. In \cite{taylor2011metric}, the metric is learned during the training process, but we do not want to learn the metric for a certain DRL model before clustering.

\subsection{Motivation and Our Contribution}
\label{subsec:contribution}
In this paper, we aim to find the initial states that can help reduce hard exploration and redundant exploration during the training process to learn robust fall recovery efficiently. To ensure robustness, the initial states need to cover as much of the state space as possible. We expect the centroids of the static-pose clusters to serve as the ideal initial states.

In terms of the metric for clustering, we propose the accessibility metric in \ref{subsec:access} to overcome the shortcomings of the Euclidean distance mentioned in \ref{subsec:related}.

For pattern-based clustering, we tried several common methods but they all failed. The directed Louvain method \cite{dugue2015directed} failed because it cannot properly distinguish the direction of the edges (state transitions in this paper) \cite{kim2010finding}. The Infomap method \cite{bohlin2014community} failed because unreachable states could exist in the same cluster, which means that the agent was expected to explore states that can hardly be reached from the initial states. The spectral clustering method \cite{gleich2006hierarchical} failed because the number of clusters was too small, which indicates low robustness. To obtain the ideal clustering effect, we propose a new accessibility-based clustering method, i.e., K-Access, in \ref{subsec:kAccess}.

Our contributions in this paper include:
\begin{itemize}
    \item An accessibility metric to quantify the level of difficulties for a robot to transition from one physical state/configuration to another;
    \item A K-Access algorithm based on the accessibility metric for state-space clustering, which is adapted from the K-Means++ algorithm;
    \item A pipeline of automatically discovering feasible initial states, based on their inter-connected transitions, for efficient learning of robot fall recovery and other tasks.
\end{itemize}

With our proposed method, the data efficiency of DRL models can be greatly improved by avoiding repeated and redundant explorations, and the robustness can be greatly enhanced because of the wide range of searched states and their explored inter-connections.

\section{Methodology}
\label{sec:methodology}
In this section, the concept of accessibility is firstly introduced in \ref{subsec:access}. The criteria of good initial state distributions are discussed in \ref{subsec:goodIni}, and the K-Access algorithm is presented in \ref{subsec:kAccess}. In \ref{subsec:fallrecL}, the DRL is applied for fall recovery, and the entire pipeline is shown in Fig. \ref{fig:pipeline}. Generic principles to apply our proposed method to learning other locomotion tasks are presented in \ref{subsec:genericPrinciple}.

\subsection{Accessibility}
\label{subsec:access}
% definition of accessibility
% choice of initial states
\begin{figure}[tpb]
  \centering
  \includegraphics[width=86mm]{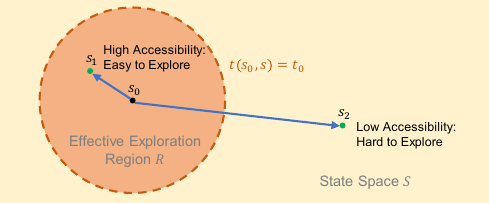}
  \caption{An illustration of accessibility. The accessibility value corresponds to the difficulty of a state being explored from the initial state.}
  \label{fig:illu_access}
    \vspace{-5mm}
\end{figure}

To model the difficulty of transitions from one state to another, we propose the accessibility metric. Figure \ref{fig:illu_access} demonstrates the concept of accessibility. Consider an initial state $s_0$ in the state space $S$. There is a region $R\subseteq S$ that can be easily and effectively explored from $s_0$, and this region can be mathematically defined as
\begin{equation}
    R = \{s|s\in S, t(s_0,s)<t_0\},
\end{equation}
where $t(s_0,s)$ is the minimal time cost (in seconds) of the transition from $s_0$ to $s$, and $t_0$ is a positive value. $R$ is called the effective exploration region of $s_0$.

Consider another two states $s_1\in R$ and $s_2\notin R$. We can say that the accessibility from $s_0$ to $s_1$ is high, and it is easy to explore $s_1$ from $s_0$. In contrast, the accessibility from $s_0$ to $s_2$ is low, and it is difficult to explore $s_2$ from $s_0$.

Mathematically, we define the accessibility from $s_i\in S$ to $s_j \in S$ as
\begin{equation}
    \text{access}(s_i,s_j) = e^{-t(s_i,s_j)}.
\end{equation}
In practice, it is difficult to get the minimal time cost $t(s_i,s_j)$, and we approximate it with PD tracking (see \ref{subsec:accValue}).
The range of accessibility is $\left[0,1\right]$.
If the accessibility is zero, $s_j$ is unreachable from $s_i$. If the accessibility is one, $s_i = s_j$.

\subsection{What Makes Good Initial States}
\label{subsec:goodIni}
\begin{figure}[tp]
  \centering
  \includegraphics[width=80mm]{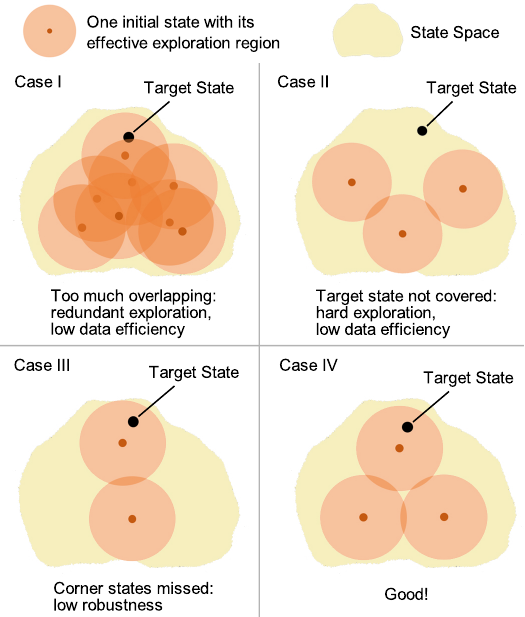}
  \caption{Different cases of initial state distributions. The target state is what the agent is expected to achieve during exploration (see \ref{subsec:genericPrinciple}). Randomized initialization corresponds to case I, and we expect the auto-discovered initial states are similar to case IV.}
  \label{fig:configs}
  \vspace{-5mm}
\end{figure}

% criteria 'good'
With the concept of accessibility, we can evaluate whether a distribution of initial states is good for the DRL exploration. Figure \ref{fig:configs} shows different cases of initial state distributions. For data efficiency, redundant exploration and hard exploration are detrimental. For robustness, the effective exploration regions should cover as much state space as possible. For the trade-off between data efficiency and robustness, we need to select the initial states and their effective exploration regions with sufficient coverage but minimal redundancy.

To ensure enough coverage, we propose to randomly sample a wide range of static poses. Then we cluster the sampled poses to reduce the redundancy. We expect that the centroids of the obtained clusters can make good initial states. Their effective exploration regions should cover most of the state space, and there should be little overlapping.

Based on the discussion above, we can evaluate whether the obtained clusters are good. 
% Figure \ref{fig:good_cluster} shows different spatial relationships between two clusters. 
If the inter-cluster accessibility is too high, there is much overlapping since the centroids can be too close to each other, leading to redundant exploration. If the intra-cluster accessibility is too low, the coverage is low since many samples are not in the effective exploration region of their centroids. Hence, for good clustering results, the inter-cluster accessibility should be low, and the intra-cluster accessibility should be high.

% comment 3.2 which gives an overview of the whole pipeline, should be moved to the top of the fourth page and use letters (a-d) for each of the four phases

\begin{figure}[t]
\vspace{-4mm}
\begin{algorithm}[H]
	\caption{K-Access$(k,A)$}
	\begin{algorithmic}[1]
\renewcommand{\algorithmiccomment}[2][.4\linewidth]{\leavevmode\hfill\makebox[#1][l]{$\lhd$~#2}}
		\STATE $cIndex\leftarrow \text{zeros}(k)$;\COMMENT{indices of centroids}
		\STATE $cIndex[0]\leftarrow \text{randInt}(0,k)$; \label{alg:sel_c0}
		\FOR{$i=1$ to $k-1$}\label{alg:new_c_begin}
		\STATE $CAccess\leftarrow \sum\limits_{j=0}^{i-1}\left(A\left[cIndex[j],:\right] + A\left[:,cIndex[j]\right]\right)$;
		\STATE $cIndex[i]\leftarrow \argmin\limits_j CAccess[j]$;\COMMENT{initialize $cIndex$}
		\ENDFOR\label{alg:new_c_end}
		\STATE $assignment[l]\leftarrow \argmax\limits_{c\in cIndex}A[c,l],\forall l<n,l\in \mathbb{N}$;\label{alg:assign1}
		\STATE $preassign\leftarrow \text{zeros}(n)$;\COMMENT{previous $assignment$}
		\WHILE{$preassign\neq assignment$}
		\STATE $preassign\leftarrow assignment$;
		\STATE $cIndex[i]\leftarrow \argmax\limits_{m\text{ s.t. }a_m= c_i}\min\limits_{a_l= c_i}A[m,l],\forall i<k,i\in \mathbb{N}$;\label{alg:update}
		\STATE $assignment[l]\leftarrow \argmax\limits_{c\in cIndex}A[c,l],\forall l<n,l\in \mathbb{N}$;\label{alg:assign2}
		\ENDWHILE
		\RETURN $cIndex,assignment$
	\end{algorithmic}
	\label{alg:k-access}
\end{algorithm}
\vspace{-10 mm}
\end{figure}

\begin{figure*}[t]
  \centering
  \includegraphics[width=176mm]{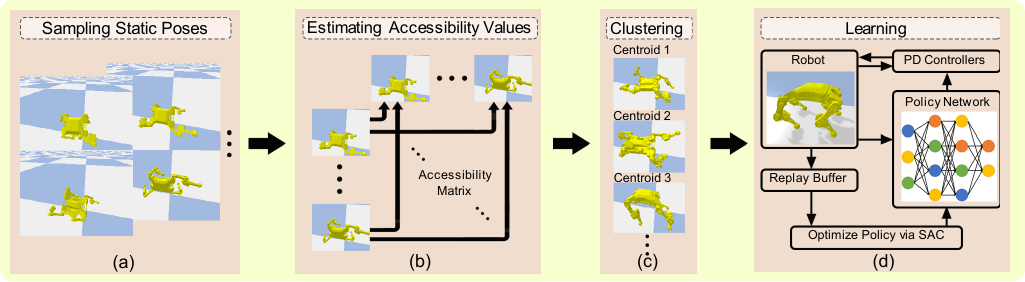}
  \caption{Pipeline of the proposed method. First, static poses are randomly sampled. Second, the estimated accessibility values are obtained via simulation. Third, the proposed K-Access algorithm selects and discovers the optimal initial states. Finally, the DRL agent learns fall recovery through exploration based on the discovered initial states.}
  \label{fig:pipeline}
  \vspace{-4mm}
\end{figure*}

\subsection{K-Access Algorithm}
\label{subsec:kAccess}
% clustering
Based on the accessibility metric proposed in \ref{subsec:access}, we refer to the K-Means++ algorithm \cite{arthur2006k} and propose the K-Access clustering method. The algorithm is presented in Algorithm \ref{alg:k-access}. The inputs are the number of clusters $k$ and the accessibility matrix $A$ for the sampled states $s_0,s_1,\dots,s_{n-1}\in S$. The size of $A$ is $(n,n)$, and $A[i,j]$ is the accessibility from $s_i$ to $s_j$ ($i,j\in \{0,\dots,n-1\}$). The outputs are the indices of the centroids $cIndex = (c_0,\dots,c_{k-1})$ and the centroids of each state's cluster $assignment=(a_0,\dots,a_{n-1})$.

Similar to the K-Means++ algorithm, the first step of K-Access algorithm is to initialize the centroids, and then we repeat the assignment step and the update step until the assignments no longer change, as described below.
\begin{enumerate}[leftmargin=*]
    \item  \textit{Initialization of centroids:} The first centroid is randomly selected (line \ref{alg:sel_c0}), and each new centroid is the furthest sample from the already selected centroids (line \ref{alg:new_c_begin}-\ref{alg:new_c_end}).
    
    \item \textit{Assignment of samples to clusters:} Each sample state is assigned to the cluster of which the centroid has the highest accessibility to this state (line \ref{alg:assign1},\ref{alg:assign2}). Note that we only consider the accessibility of single direction here.
    
    \item \textit{Update of the choice of centroids:} The new centroid has the maximal neighborhood accessibility. The neighborhood accessibility of one state is defined as the minimal accessibility value from the state to its neighbors in the same cluster (line \ref{alg:update}). To ensure robustness, we do not take the average, which differs from K-Means++.
\end{enumerate}

Finally, the K-Access algorithm can converge to a result that prefers high intra-cluster accessibility and low inter-cluster accessibility. To determine the number of clusters, we propose an index based on the discussions in \ref{subsec:goodIni} that good clustering results should have high intra-cluster accessibility and low inter-cluster accessibility. The index $I$ is defined as
\begin{equation}
    I =\overline{\log(A_{\rm intra})} - \overline{\log(A_{\rm inter})} - \alpha \cdot | \Lambda |,
\end{equation}
where $A_{intra}$ is the intra-cluster accessibility array of size $k$, $A_{inter}$ is the inter-cluster accessibility of size $(k,k)$, and $\alpha\cdot |\Lambda |$ is the regularization term. The clustering results are better if the index is larger, and the inter/intra-cluster metrics are also defined in a different way from K-Means++. To be specific, 
\begin{equation}
    A_{\rm intra}[i] = \min_{a_l= c_i, l<n}A[c_i,l],\forall i<k,
\end{equation}
\begin{equation}
    A_{\rm inter}[i,j] = 
    \left\{
\begin{aligned}
&\underset{a_l=c_i}{\text{mean }}A[a_l,c_j],\forall i\neq j,\\
& 1 ,\forall i=j,
\end{aligned}
\right.
\end{equation}
where $i<k,i\in \mathbb{N},j<k,j\in \mathbb{N}$,
\begin{equation}
    \Lambda = \left\{c_i\mid i<k,i\in\mathbb{N},\lvert\{l\mid a_l=c_i,l<n,l\in\mathbb{N}\}\rvert=1\right\},
\end{equation}
and $\alpha$ is a real value (recommended value: $1$). We penalize the number of one-sample clusters because such clusters can hardly be encountered and they may exist because of extreme circumstances. In most cases, one-sample clusters are also accessible from other clusters. Therefore, one-sample clusters should be rare or non-existent, otherwise the training process may suffer from unnecessary or redundant explorations.
In Fig. \ref{fig:pipeline} (c), the K-Access algorithm is applied for clustering, after the estimation of accessibility values for the sampled poses.

\subsection{Fall Recovery Learning}
\label{subsec:fallrecL}
\begin{figure}[t]
  \centering
  \includegraphics[width=86mm]{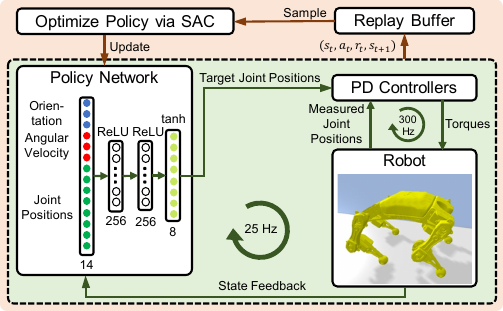}
  \caption{DRL framework overview. The policy network generates the target joint positions at 25 Hz. The actuators follow the 300 Hz impedance control based on the target and the measured joint positions. The SAC algorithm is applied for learning the feedback control policy.}
  \label{fig:system}
  \vspace{-5mm}
\end{figure}

% model & environment setup
Our DRL framework is shown in Fig. \ref{fig:system}. The state space consists of the orientation (represented by the normalized gravity vector), the angular velocity of the body, and the joint positions. Here we adopt the positional control for high learning efficiency and performance according to \cite{peng2017learning}. The outputs of the policy network are the target joint positions which update at 25 Hz. The PD controllers generate torques based on the target joint positions and the measured joint positions at 300 Hz. The SAC algorithm is applied for learning. In our implementation, we use the 8-DoF quadrupedal robot \textit{Bittle} \cite{bittle} and the PyBullet \cite{coumans2021} simulation environment.

The \textit{Bittle} is equipped with an IMU on its body. The angular velocity can be directly accessed from the IMU. The orientation is represented as the gravity vector in the body frame which is normalized to be of length $1$. The gravity vector can be computed using the IMU measurements.

\begin{table}[tp]
\setlength{\abovecaptionskip}{-0.1cm}
\caption{Reward terms for DRL.}
\label{tab:reward}
\begin{center}
\begin{tabular}{|ll|}
\hline
\multicolumn{2}{|c|}{Symbols}\\ \hline
$h$ &\hspace*{-0.6in} body height\\
$g_{\rm ori}$ &\hspace*{-0.6in} normalized gravity vector\\
$\Omega$ &\hspace*{-0.6in} body angular velocity\\
$\tau$ &\hspace*{-0.6in} vector of joint torques\\
$\omega$ &\hspace*{-0.6in} vector of joint velocity\\
$c_{\rm b}$ &\hspace*{-0.6in} 0 if the body touches the ground, otherwise 1\\
$c_{\rm f}$ &\hspace*{-0.6in} 0.3$\times$the number of feet that touch the ground\\
$h_{\rm f}$ &\hspace*{-0.6in} vector of the distances from four feet to the ground\\
$d_{\rm s}$ &\hspace*{-0.6in} the shortest distance from the robot to the ground\\
$p$ &\hspace*{-0.6in} vector of the measured joint positions\\
$p^{\rm t}$&\hspace*{-0.6in} vector of the target joint positions\\
$\hat{\cdot}$ &\hspace*{-0.6in} target value \\ \hline
\multicolumn{2}{|c|}{Reward Terms}\\ \hline
Height&$0.667\times\text{RBF}(h,\hat{h},-2000)$\\
Orientation&$0.333\times\text{RBF}(g_{\rm ori},[0,0,-1]^T,-5)$\\
Angular velocity&$0.067\times\text{RBF}(\Omega,0,-0.05)$\\
Joint torques&$0.067\times\text{RBF}(\tau,0,-5)$\\
Joint velocity&$0.067\times\text{RBF}(\omega,0,-0.05)$\\
Contact&$0.067\times c_{\rm b}+0.033\times c_{\rm f}$\\
Foot lift&$0.067\times\text{RBF}(h_{\rm f},0,-100)$\\
Jump regularization&$0.033\times\text{RBF}(d_{\rm s},0,-100)$\\
Action difference&$0.033\times\text{RBF}(p^{\rm t},p,-1)$\\
\hline
\end{tabular}
\end{center}
\vspace{-3em}
\end{table}

The reward function is the sum of reward terms in Table \ref{tab:reward}. The \textit{jump} regularization term aims to penalize the robot for actions (e.g., spinning) to adjust the orientation in the air, and the action difference term serves to reduce unrealistic large movements. We assign larger weights to the reward terms related to body height and orientation since we characterize a standing pose mainly by the body height and the orientation for the fall recovery task. The radial basis function (RBF) applied in these terms is defined as:
\begin{equation}
    \text{RBF}(x,y,\alpha)=\exp\left({\alpha\cdot \left\|x-y\right\|_2^2}\right).
\end{equation}

The learning process is the last stage (d) of the pipeline in Fig. \ref{fig:pipeline}, where we apply the centroids in (c) as the initial states for DRL of quadruped fall recovery.

\subsection{Learning Other Tasks}
\label{subsec:genericPrinciple}
The proposed method is not limited to fall recovery learning. In \ref{subsec:backflip}, we also validated our method in backflip learning. The generic principles to apply the proposed method are:
\begin{enumerate}[leftmargin=*]
    \item Perform clustering in a subspace with feasible and necessary dimensions;
    \item Ensure that the target state has the maximum expected return in the subspace;
    \item Estimate the accessibility values with time-accuracy trade-off.
\end{enumerate}

The first principle is about how we define the subspace for clustering. Some dimensions such as velocities are not suitable for estimating the accessibility values.  Also, necessary dimensions to identify the initial states must be included.

The second principle is about the reward function and the target state. In fall recovery, the target state corresponds to the standing pose. In other tasks such as backflipping, the target state can refer to a good starting pose that can get the maximum episode reward in the future.

The third principle is about the estimation of accessibility values. In fall recovery, we apply the response time of PD control. However, there can be errors in some cases, e.g., when large torques make the robot fly. Also, more complicated and high-level controllers can be necessary for other scenarios. In such cases, a trade-off between complexity and estimation accuracy needs to be considered.

\section{Implementation and Results}
\label{sec:implementation}
%In this section, we present the implementation of our pipeline in detail.

\subsection{Sampling Static Poses}
\label{subsec:poseSample}
In the PyBullet environment, we randomly initialized the \textit{Bittle} robot with roll angle $\phi \sim \mathrm{U}(-\pi,\pi)$, pitch angle $\theta \sim \mathrm{U}(-\frac{\pi}{2},\frac{\pi}{2})$, yaw angle $\psi = 0$, and joint positions $p \sim \mathrm{U}(-\frac{5}{6}\pi,\frac{5}{6}\pi)$. Self-collision cases were abandoned. 

The robot was dropped from 0.35 m above the ground. If the robot was stationary within 2 seconds (with negligible velocity, angular velocity, and distance to the ground), we recorded the final joint positions, the final roll angle, the final pitch angle, and the body height. We assume the ground to be flat for sampling.

With $12\times$ multiprocessing, we sampled 2.4k static poses within an hour on an ordinary desktop machine.

\subsection{Estimating Accessibility Values}
\label{subsec:accValue}
To estimate the accessibility from static pose A to static pose B, we initialized the robot with static pose A. Then we sent a command to the PD controllers with the joint positions of B as the target positions. If the robot entered static equilibrium within 3 seconds, we checked whether the state of robot was close to the state of B. If these two states were close enough, we recorded the time cost $t$ and the estimated accessibility from A to B was $e^{-t}$. Otherwise, we set the accessibility to be $10^{-8}$ instead of zero because all the sampled static poses are assumed to be accessible.

We randomly selected 1k samples from the sampled static poses. With $12\times$ multiprocessing, we obtained the 1M accessibility values for the 1k$\times$1k accessibility matrix within 20 hours on an ordinary desktop machine.

\subsection{Clustering}
\label{subsec:clustering}

\begin{figure}[t]
  \centering
  \vstretch{0.92}{\includegraphics[width=86mm]{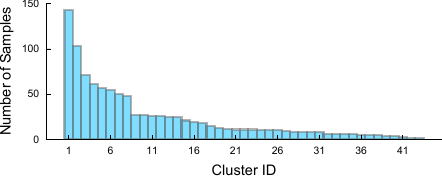}}
  \caption{Number of samples within each cluster when there are $k=43$ clusters. The clusters are sorted by the number of samples assigned to them.}
  \label{fig:num_samples}
  \vspace{-2mm}
\end{figure}

\begin{figure}[t]
  \centering
  \includegraphics[width=80mm]{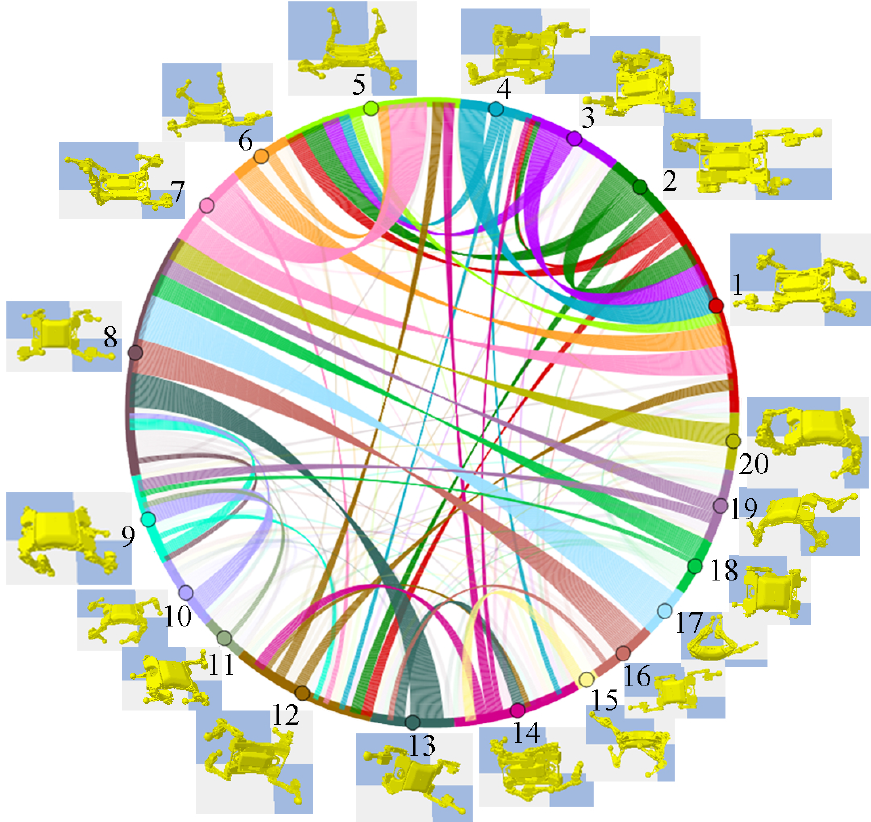}
  \caption{Visualization of the inter-cluster accessibility of top-20 clusters. Inter-cluster accessibility values above 0.15 are highlighted, and those values below 0.05 are omitted here for clarity.}
  \label{fig:vis_inter}
  \vspace{-5mm}
\end{figure}

\begin{figure*}[t]
    \centering
    \includegraphics[width=174mm]{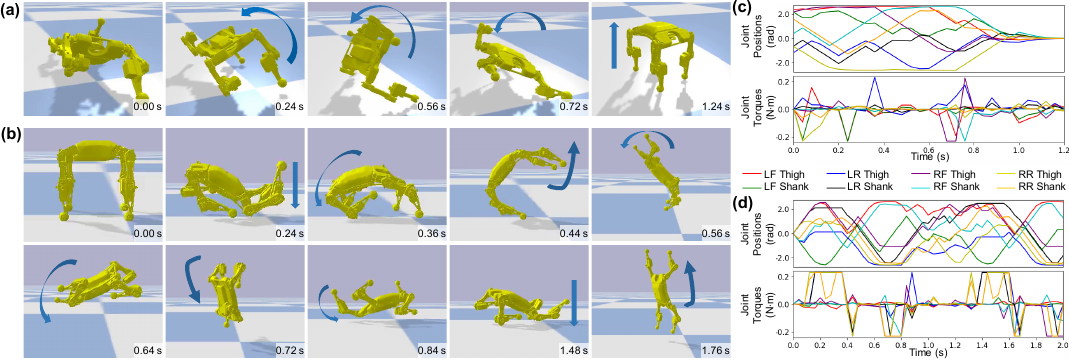}
    \caption{Snapshots of (a) fall recovery and (b) continuous backflips. (c) shows the joint positions and torques for (a), and (d) shows those for (b).
    }
    \label{fig:snapshot}
  \vspace{-5mm}
\end{figure*}

We applied the proposed K-Access algorithm to the 1k$\times$1k accessibility matrix obtained in \ref{subsec:accValue}. The $\alpha$ value for the index is 1. To determine the number of clusters $k$, we tried different $k$ values and obtained the maximum index value when $k=43$. The number of samples in each of the 43 clusters is shown in Fig. \ref{fig:num_samples}, indicating that the static poses of the robot are subjected to a long-tail distribution.

Figure \ref{fig:vis_inter} visualizes the inter-cluster accessibility of top-20 clusters via chord diagram, which can also contribute to pose taxonomy analysis \cite{borras2015whole} \cite{borras2017whole}. Different clusters correspond to different contact cases, and the inter-cluster accessibility corresponds to the difficulty of transitions.

\subsection{Deep Reinforcement Learning}
\label{subsec:training}

We applied six kinds of initial state distributions:
\begin{enumerate}[leftmargin=*]
    \item Centroids of the obtained 43 clusters by K-Access (abbreviated to KA);
    \item Centroids of the obtained 33 clusters by K-Means++ based on the generalized Dunn's index $\nu 43$ \cite{bezdek1995cluster} (abbreviated to KM);
    \item Centroids of the obtained 14 clusters by weighted K-Means++ based on $\nu 43$ (abbreviated to WKM, gravity vector weighted by 2);
    \item Nine initial poses applied in \cite{yang2020multi} (abbreviated to 9-Pose);
    \item One lying pose (abbreviated to 1-Pose);
    \item Random static poses (abbreviated to RND).
\end{enumerate}

\begin{figure}[t]
  \centering
  \vstretch{0.95}{\includegraphics[width=84mm]{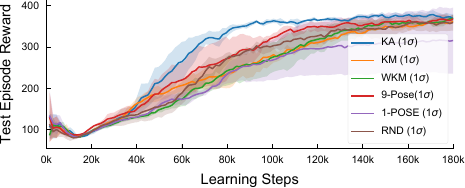}}
  \caption{Learning curves of fall recovery for different initial state distributions. The proposed method (KA) shows the highest data efficiency. There are 300 steps per episode, the discount factor is 0.987. Averaged over 3 random seeds, with $\pm$1 SD. in shadow.}
  \label{fig:learning_curve}
  \vspace{-6mm}
\end{figure}

We demonstrate the proposed method can greatly improve the data efficiency, based on the learning curves for fall recovery using the same test samples shown in Fig. \ref{fig:learning_curve}. 
% The centroids of KA, KM, and WKM are also visualized in Fig. \ref{fig:layout} to validate our hypothesis in \ref{sec:methodology}.

% \begin{figure}[t]
%     \centering
%     \includegraphics[width=88mm]{figs/fig10.pdf}
%     \caption{Visualization of the obtained clusters for overlapping and coverage. The red points are the target standing pose. The blue points are the centroids of top 20 clusters for (a)-(b), and all of the 14 clusters for (c). The positions of the centroids follow a spring layout based on the inter-cluster accessibility values. The average accessibility values from the centroids to their neighbors, representing the expected burden of exploring the centroids, determine the radii of the circles. The K-Access centroids generate better initial states according to Fig. \ref{fig:configs}.}
%     \label{fig:layout}
%     \vspace{-2mm}
% \end{figure}

For robustness, agents of the best seeds were tested on another 500 static poses. There are 75 steps (3 seconds) per episode during the test. The performance scores are presented in Table \ref{tab:score}. It can be seen that the proposed method can be used to learn robust fall recovery policies with 25\% fewer steps than other distributions. Test runs are shown in the accompanying video. Snapshots are in Fig. \ref{fig:snapshot}(a).

\subsection{Backflip Learning}
\label{subsec:backflip}

\begin{figure}[t]
    \centering
    \vstretch{0.95}{\includegraphics[width=84mm]{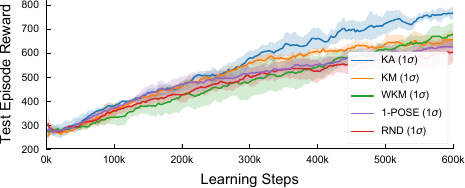}}
    \caption{Learning curves of backflipping for different initial state distributions. Averaged over 3 random seeds, with $\pm$1 SD. in shadow.}
    \label{fig:learning_curve_backflip}
    \vspace{-2mm}
\end{figure}

We applied our proposed method to learn backflipping and the learning curves using the same test samples are in Fig. \ref{fig:learning_curve_backflip}. Here we only clustered the static poses with roll angle less than $60 \deg$. See snapshots in Fig. \ref{fig:snapshot}(b). Test runs and learning details are in the accompanying video.

% The visualization of inter-cluster accessibility also indicates that the patterns can be heuristic for learning fall recovery motions. Moreover, such patterns can also help design strategies for optimization-based methods \cite{castano2019design}.

% For robustness over a large range of initial states, there are also other methods, such as region of attraction (RoA) expansion \cite{kumar2018improving} \cite{borno2017domain}. In \cite{kumar2018improving}, the learning process is also expedited by adaptive sampling, but the technique is based on the simplified representation of RoA that is difficult to implement in a high-dimensional nonlinear state space. It is also hard to combine the RoA expansion with the DRL process, since the DRL policy does not explicitly bridge the intermediate states as in \cite{borno2017domain}. However, the RoA expansion can be applied after the learning process to improve the robustness, and the robustness during the DRL process within finite steps can reduce the time cost and complexity of the subsequent RoA expansion.
\begin{table}[t]
\setlength{\abovecaptionskip}{-0.25cm}
\caption{Performance on the Test Poses}
\label{tab:score}
\begin{center}
\begin{tabular}{ccccc} 
\hline
\multirow{2}*{\shortstack{\textbf{Initial}\\\textbf{States}}} &
\multirow{2}*{\shortstack{\textbf{Training}\\\textbf{Episodes}}} &
\multicolumn{2}{c}{\textbf{Episode Reward}} & 
\multirow{2}*{\shortstack{\textbf{Success Rate}\\\textbf{in 3 s (\%)}}} \\ \cline{3-4}
&& \textbf{Mean} & \textbf{SD.} &     \\ \hline 
{KA (proposed)} &\textbf{1200}&\textbf{81.93}&\textbf{11.84}&\textbf{99.4} \\
{KM} &1600&74.84&20.14& 91.8\\
{WKM} &1600&79.04&18.20&94.4 \\
{9-Pose} &1600&73.35&17.01& 91.6\\
{1-Pose} &1600&73.09&17.75&92.6 \\
{RND} &1600&79.97&17.32&94.6 \\ \hline 
\end{tabular}
\end{center}
\vspace{-2em}
\end{table}

\section{Conclusion}
\label{sec:conclusion}
\label{sec:discussion}
In this paper, we propose to automatically discover initial states for DRL of locomotion skills via the accessibility metric and the K-Access clustering algorithm. With the obtained centroids as the initial states, the data efficiency can be greatly improved, and the robustness can be guaranteed. Despite the extra computation to estimate accessibility values before clustering, the centroids are only computed once based on this learning-free metric, and they can boost the training process regardless of the choices of DRL algorithms, hyperparameters, and reward functions.

The K-Access algorithm performs better than the existing clustering methods because: 1) the direction of transitions is taken into consideration; 2) the neighbors in one cluster are all easy to explore from the centroid; 3) the number of clusters can be assigned and determined by the index value. 
The generalization to backflipping also shows that intermediate and unstable poses can also be better explored with our method.

% Compared to the Euclidean distance metric, the proposed accessibility metric does not suffer from the non-orthogonality of the feature space and the undirected distances. Therefore, it can model the difficulty of transitions from one state to another much better. 

% % With high intra-cluster accessibility and low inter-cluster accessibility, the centroids obtained by the K-Access algorithm can make good initial states.

% Compared to random initialization, our method is more data efficient, as shown by our results in Fig. \ref{fig:learning_curve} and Fig. \ref{fig:learning_curve_backflip}. Compared to the manually predefined initialization, our method is more robust, as benchmarked in Table \ref{tab:score}. 

Future work will focus on the possible extensions to other tasks that can benefit from the proposed technique. We will also do hardware validation and try other methods for better accessibility estimation, e.g., neural networks.

%%%%%%%%%%%%%%%%%%%%%%%%%%%%%%%%%%%%%%%%%%%%%%%%%%%%%%%%%%%%%%%%%%%%%%%%%%%%%%%%
% \addtolength{\textheight}{-0cm}   % This command serves to balance the column lengths
                                  
% \label{sec:appendix}

% \section*{ACKNOWLEDGMENT}
% \label{sec:acknowledge}
% This is the acknowledgment.

%%%%%%%%%%%%%%%%%%%%%%%%%%%%%%%%%%%%%%%%%%%%%%%%%%%%%%%%%%%%%%%%%%%%%%%%%%%%%%%%
% References are important to the reader; therefore, each citation must be complete and correct. If at all possible, references should be commonly available publications.  
\bibliographystyle{IEEEtran}
\balance
\bibliography{IEEEabrv,References}

\end{document}